\documentclass[journal,twoside,web]{ieeecolor}
\usepackage{tmi}
\usepackage{cite}
\usepackage{amsmath,amssymb,amsfonts}
\usepackage{algorithmic}
\usepackage{graphicx}
\usepackage{textcomp}
\usepackage{multirow, multicol}
\usepackage[colorlinks=true,linkcolor=black,citecolor=black,urlcolor=blue]{hyperref}
\usepackage{booktabs}
\usepackage[draft]{changes}
\definechangesauthor[name={Matias}, color=blue]{MC}

\let\labelindent\relax
\usepackage{enumitem}
\usepackage{comment}
\usepackage{etoolbox}
\usepackage{url}


\newtheorem{theorem}{Theorem}
\makeatletter
\def\@begintheorem#1#2{\trivlist
  \item[\hskip \labelsep \bfseries #1\ #2.]\normalfont}
\makeatother

\def\BibTeX{{\rm B\kern-.05em{\sc i\kern-.025em b}\kern-.08em
T\kern-.1667em\lower.7ex\hbox{E}\kern-.125emX}}
\begin{document}
\title{ConfIC-RCA: Statistically Grounded Efficient Estimation of Segmentation Quality}

\author{Matias Cosarinsky, Ramiro Billot, Lucas Mansilla, Gabriel Jimenez, Nicol\'as Gaggion, Guanghui Fu, \\Tom Tirer, \IEEEmembership{Member, IEEE}, and Enzo Ferrante
\thanks{Matias Cosarinsky, and Enzo Ferrante are with the Institute of Computer Sciences (CONICET - Universidad de Buenos Aires), Buenos Aires, Argentina (emails: mcosarinsky@dc.uba.ar, eferrante@sinc.unl.edu.ar).}
\thanks{Nicol\'as Gaggion is with APOLO Biotech, Buenos Aires, Argentina, and with the Institute of Computer Sciences (CONICET - Universidad de Buenos Aires), Buenos Aires, Argentina (e-mail: ngaggion@sinc.unl.edu.ar).}
\thanks{Ramiro Billot is with Universidad Nacional de San Martin, San Martin, Argentina (email: rbillot@estudiantes.unsam.edu.ar).}
\thanks{Tom Tirer is with the Faculty of Engineering, Bar-Ilan University, Ramat Gan, Israel (e-mail: tirer.tom@gmail.com).}
\thanks{Lucas Mansilla is with the Research Institute of Signals, Systems and Computational Intelligence, sinc(i) (CONICET - Universidad Nacional del Litoral), Santa Fe, Argentina (e-mail: lmansilla@sinc.unl.edu.ar).}
\thanks{Gabriel Jimenez and Guanghui Fu are with Sorbonne Université, Institut du Cerveau - Paris Brain Institute - ICM, CNRS, Inria, Inserm, AP-HP, Hôpital de la Pitié Salpêtrière, Paris, France (emails: gabriel.jimenez@icm-institute.org, guanghui.fu@icm-institute.org).}
\thanks{EF was supported by the Google Award for Inclusion Research program and a Googler Initiated Grant.}
\thanks{The work of TT was supported by the Israel Science Foundation (No. 1940/23) and MOST (No. 0007091) grants.}}

\maketitle              
\begin{abstract}
Assessing the quality of automatic image segmentation is crucial in clinical practice, but often very challenging due to the limited availability of ground truth annotations. Reverse Classification Accuracy (RCA) is an approach that estimates the quality of new predictions on unseen samples by training a segmenter on those predictions, and then evaluating it against existing annotated images. In this work we introduce ConfIC-RCA (Conformal In-Context RCA), a novel method for automatically estimating segmentation quality with statistical guarantees in the absence of ground-truth annotations, which consists of two main innovations. First, In-Context RCA, which leverages recent in-context learning models for image segmentation and incorporates retrieval-augmentation techniques to select the most relevant reference images. This approach enables efficient quality estimation with minimal reference data while avoiding the need of training additional models. Second, we introduce Conformal RCA, which extends both the original RCA framework and In-Context RCA to go beyond point estimation. Using tools from split conformal prediction, Conformal RCA produces prediction intervals for segmentation quality providing statistical guarantees that the true score lies within the estimated interval with a user-specified probability. Validated across 10 different medical imaging tasks in various organs and modalities, our methods demonstrate robust performance and computational efficiency, offering a promising solution for automated quality control in clinical workflows, where fast and reliable segmentation assessment is essential. The code is available at \href{https://github.com/mcosarinsky/Conformal-In-Context-RCA}{https://github.com/mcosarinsky/Conformal-In-Context-RCA}.

\end{abstract}


\section{Introduction}
\IEEEPARstart{I}{mage} segmentation plays a crucial role in many medical practices, particularly for treatment planning, clinical decision making and monitoring disease progression. In these scenarios, evaluating the performance of segmentation methods becomes of utmost importance. Typically, the evaluation is performed on annotated databases using metrics that measure the agreement between predicted segmentations and \emph{ground truth} (GT) labels. Commonly used metrics include the Dice-Sørensen coefficient (DSC) and other overlap-based measures \cite{crum2006generalized}, as well as distance-based metrics, such as the Hausdorff distance or the average symmetric surface distance (ASSD) \cite{evaluation-metrics,taha2015metrics}. 

Once a segmentation method is deployed in practice, continuously assessing its real performance on new unseen data samples becomes a challenging task, mainly due to the limited availability of reference annotations in clinical settings. However, evaluation remains a critical task, especially for identifying instances where an image segmentation method may fail in automatic pipelines due to unexpected changes, e.g. in patient demographics\cite{gaggion2023unsupervised} or acquisition protocols. This motivates the development of methods that can automatically estimate segmentation quality when no ground truth is available. 

Reverse Classification Accuracy (RCA) \cite{rca} is a framework originally designed to predict the performance of an image segmentation method in scenarios where GT masks are unavailable. It works by using the predicted segmentation of a new image as a pseudo-ground truth to train a reverse classifier (a segmentation network in our case)\footnote{Note that we use the term \emph{reverse classifier} as introduced in the original publication \cite{rca}, although it would be more accurately described as a \emph{reverse segmenter}. We will use both terms interchangeably.}. This model is then applied to a set of reference images with known GT, and its performance is evaluated using standard metrics like the DSC. The distribution of scores over the reference set serves as a proxy for the quality of the original segmentation. Even though RCA has been widely adopted in the medical imaging community, it still suffers from two main limitations: it is very inefficient in computational terms and it lacks a mechanism to quantify the uncertainty of its predictions. In this work, we address both limitations by introducing \textbf{ConfIC-RCA (Conformal In-Context RCA)}, a \emph{model-agnostic framework} that combines two complementary advances.\\

At the core of our approach is \textbf{Conformal RCA}, which leverages \emph{split conformal prediction} \mbox{\cite{algorithmic-learning,conformalprediction,cqr}} to produce \emph{statistically grounded confidence intervals} for segmentation quality scores. Unlike traditional RCA, which provides only a single point estimate (e.g., mean or maximum over reference scores), Conformal RCA quantifies the uncertainty associated with each prediction and provides statistical guarantees that the true segmentation quality lies within the predicted interval with a user-specified probability. These guarantees are distribution-free and hold under the standard exchangeability assumption between calibration and test samples. This allows for a more informative and reliable assessment of segmentation quality at the individual image level, directly addressing a major limitation of existing methods.

To improve practicality and efficiency, we incorporate \textbf{In-Context RCA}, which uses in-context learning segmentation models \mbox{\cite{wang2023seggpt,universeg,medicalsam2}} that can perform new segmentation tasks by conditioning on a small set of labeled examples without requiring any additional training. In practice, this means we can segment a new image by providing only a few image-mask pairs, avoiding the need to train a new segmentation model. By using in-context segmentation models as reverse classifiers, our framework significantly reduces the computational cost of RCA while maintaining high accuracy, making it more suitable for clinical use.  

\section{Related work}

\subsection{Automatic Quality Control and InContext Learning} Despite its relevance, few approaches have addressed automatic quality control (QC) of segmentations thus far. Many previous works have relied on extracting features from segmentations and fitting a model to these features. Such approaches have been applied to prostate segmentation \cite{sunoqrot2020quality}, cardiac segmentation \cite{xu2009automated}, and a more generic framework  has been used to assess the quality for automated segmentations for lung and liver in \cite{kohlberger2012evaluating}. However, these methods are constrained by the feature extraction, which can limit their adaptability and generalizability to diverse clinical data. 

Deep generative and learning-based methods have been proposed to improve QC. For instance, \cite{deep_generative-QC} train a generative model to learn a manifold of good-quality cardiac MRI segmentations and then evaluate new segmentations by their projection error. Similarly, \cite{liu-shape_model} use a shape-model-based VAE to capture a low-dimensional feature space of acceptable segmentations and predict quality via a learned regressor. \cite{zuluaga} propose global and pixel-wise quality measures using a convolutional autoencoder trained on ground-truth masks. QCResUNet \cite{qcresunet} jointly predicts subject-level quality scores and voxel-level error maps using a multi-task deep network. While these methods achieved strong performance, they often require training specialized models per task or heavy computation.

Bayesian inference methods have also been proposed. For instance, \cite{audelan2019unsupervised} follow an unsupervised Bayesian framework for brain tumor and left ventricle myocardium segmentation, measuring segmentation adequacy via a probabilistic model combining intensity and spatial priors. Such methods are principled but computationally demanding and usually tailored to specific anatomical targets. Another class of QC approaches derives quality estimates from model uncertainty. A common strategy is to compute epistemic uncertainty maps from multiple segmentation outputs using techniques such as MC Dropout or test‑time augmentation (TTA) \cite{gal2016dropoutbayesianapproximationrepresenting, tta, fuzzy-uncertainty}.  Summary statistics such as entropy or variance over these maps can serve as image‑level indicators of segmentation uncertainty or quality \cite{uncertainty-maps}. While attractive for not requiring additional training, these uncertainty-based metrics can show inconsistent correlation with true segmentation quality.

Additionally, a CNN-based approach was developed for automated quality control of left ventricle segmentation in cardiac MRI \cite{fournel2021medical}. In spite of demonstrating the potential of deep learning for automatic quality assessment, this method requires generating a synthetic dataset and further training multiple CNN models.

Other previous studies rely on a reverse-testing strategy, where a new classifier is trained on predictions on the test data and evaluated again on the training data \cite{reverse-testing,reverse-validation,chexmask}. The limitation of these approaches is that they only provide an estimate of the average performance rather than case-specific estimations. Building on this idea, Reverse Classification Accuracy (RCA) has been shown to provide accurate predictions of segmentation quality on a per-image basis \cite{rca,robinson2017automatic}. In \cite{rca}, Atlas-based segmentation has shown to be the most effective reverse classifier for medical image segmentation. 
However, the computational time required to complete atlas based segmentation on a single instance is a limitation, making it impractical for real-time quality control. In \cite{robinson2018real} the authors propose using a neural network to directly estimate prediction quality, aiming to mitigate the time constraint. However, this approach necessitates training a separate CNN for each task, which significantly limits its practicality and scalability. More importantly, both methods \cite{rca,robinson2017automatic}  have only been used for anatomical segmentation and not for lesions, which are usually more challenging  due to the lack of regularity.

Recently, Few-shot Semantic Segmentation (FSS) methods ---which extend in-context learning to address segmentation tasks--- have shown great potential in the medical domain, particularly with limited annotations \cite{ouyang2022self}. While traditional approaches are often based on prototypical networks \cite{snell2017prototypicalnetworksfewshotlearning,ouyang2020self}, more advanced models such as UniverSeg \cite{universeg} and foundational models like SAM 2 \cite{sam2,fsmedsam2,medicalsam2} delivered more promising results on a wide range of medical tasks. This enhanced generalizability is largely due to their training on vast amounts of diverse data. Here we will use FSS to improve RCA performance and efficiency.

\begin{figure*}[t]
    \centering
    \includegraphics[width=0.7\textwidth]{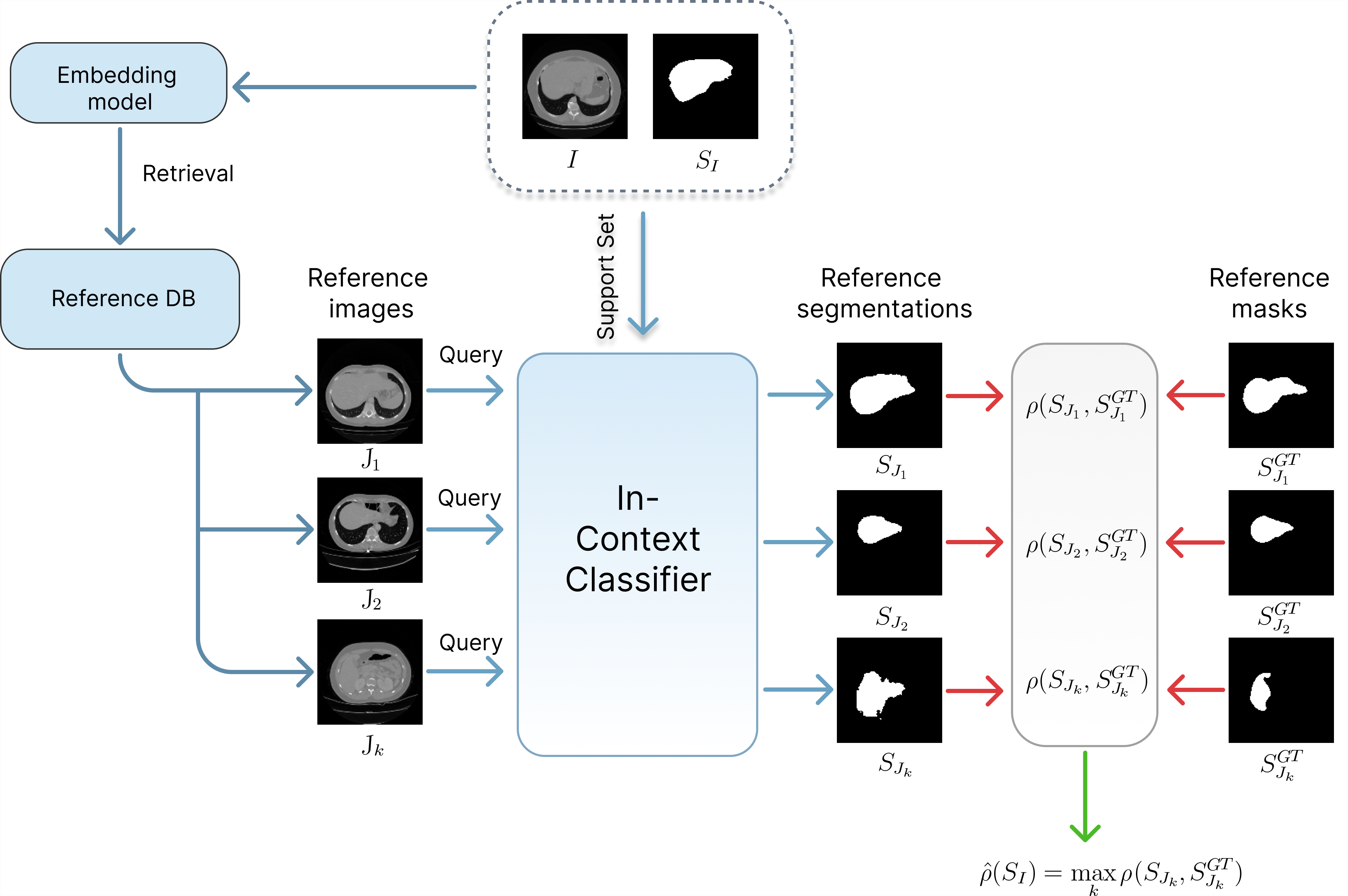} 
    \caption{General outline of the In-Context RCA framework. The quality of a predicted segmentation $S_I$ on an image $I$ is estimated by leveraging $(I, S_I)$ as the support set of an In-Context Classifier or segmenter. This classifier is then applied to segment a reference dataset, selected through retrieval augmentation. Finally, the best segmentation result based on an evaluation metric $\rho$, is used to predict the quality of $S_I$ following Eq.~\eqref{eq:RCA-estimate}.}
    \label{fig:framework}
\end{figure*}

\subsection{Performance Interval Estimation} 
While most existing methods for segmentation quality estimation focus on predicting point estimates, few works aim to quantify the uncertainty of such estimations. Recently, the idea of estimating performance ranges has emerged as a promising direction. One notable approach is the method proposed in \cite{range-prediction}, which applies split conformal prediction to produce intervals for segmentation quality scores with formal coverage guarantees. 
To do so, however, they must either train specialised probabilistic segmentation models that can sample multiple plausible outputs for the same input---such as Probabilistic U‑Net \cite{probabilisticunet} or PHiSeg \cite{phiseg}---or rely on alternative uncertainty estimation techniques like test‑time augmentation \cite{tta}, Monte‑Carlo dropout \cite{gal2016dropoutbayesianapproximationrepresenting}, or ensembling \cite{deepensembles}; all of which add further implementation overhead and yielded lower performance in the reported experiments. In addition, the method constructs its intervals from nonconformity scores based on the softmax scores \cite{li2022estimatingmodelperformancedomain,pmlr-v250-kohler24a}, which are known to suffer from calibration issues \cite{guo2017calibrationmodernneuralnetworks,dabah2024temperature}. In this work, we introduce Conformal RCA, which unlike previous approaches, is model-agnostic and can be used to estimate quality intervals for segmentation masks regardless of the underlying architecture or training procedure used to produce them. Unlike \cite{range-prediction}, we base our conformal prediction framework on Conformalized Quantile Regression (CQR), leveraging its ability to produce tighter prediction intervals than alternative conformal approaches commonly used for regression \cite{cqr}, and introducing novel modifications tailored to the RCA procedure.\\

\vspace{0.5em}
\noindent \textbf{Our Contributions}:
\begin{itemize}[label=\textbullet]
    \item We introduce \textbf{ConfIC-RCA}, the first model-agnostic framework for segmentation quality estimation that provides prediction intervals with statistical guarantees, enabling image-level uncertainty quantification. We base our approach on CQR, but unlike prior work that trains a quantile regression model, we estimate the quantiles directly from the RCA reference set, avoiding additional training.
    \item We propose \textbf{In-Context RCA} as a practical enhancement of the traditional RCA framework, leveraging few-shot segmentation models to improve computational efficiency and adapt to new tasks without requiring additional training.
    \item We introduce Retrieval-Augmented RCA, a more efficient variant of RCA that uses embedding-based retrieval to select the most relevant reference images to be used as few-shot examples, enhancing efficiency and accuracy.
    \item We conduct extensive experiments on 10 diverse medical imaging datasets, demonstrating strong performance across different modalities and anatomy. We significantly broaden the scope of RCA while reducing inference time.
\end{itemize}

\section{In-Context RCA}

Suppose we deployed to production an X-ray lung segmentation model, and we want to monitor the quality of the segmentation $S_I$ produced for a new image $I$. Let us also assume access to a reference database $D = \{(J_i, S_{J_i}^{GT})\}_{i=1}^m$ of annotated images (in our case, X-ray images with expert annotations). The core idea behind RCA is to assess the quality of a segmentation $S_I$ corresponding to a new image $I$ by training a \emph{reverse segmenter} $f_{I, S_I}$ (or reverse classifier) using \textit{only} $S_I$ as pseudo-ground truth for training. The reverse segmenter is then applied to each image in $D$, generating segmentations $f_{I, S_I}(J_i) = S_{J_i}$. The quality of these segmentations is then assessed by computing a chosen quality metric $\rho$ (e.g DSC) for each $S_{J_i}$ and its corresponding ground truth $S_{J_i}^{GT}$. This process yields a set of scores:
\begin{equation}
    \mathcal{Y} = \left\{ y^{(k)}(S_I) = \rho(S_{J_k}, S_{J_k}^{GT})  \right\}_{k=1}^m.
    \label{eq:RCA-scores}
\end{equation}
Traditionally, RCA collapses this set into a point estimate of segmentation quality by taking the best metric value (or the mean) over $\mathcal{Y}$:
\begin{equation}
    \hat{\rho}(S_I) = \max_k \rho(S_{J_k}, S_{J_k}^{GT}),
    \label{eq:RCA-estimate}
\end{equation}   
assuming that higher values of $\rho$ correspond to better quality estimations. The underlying hypothesis is that if $S_I$ is of high quality, the RCA segmenter should perform well on at least some of the reference images, resulting in high values of $\rho$. Conversely, if $S_I$ is of poor quality, the segmenter is expected to perform poorly on the reference database. 

In this work, we propose to adopt in-context segmentation models as reverse segmenters, resulting in In-Context RCA (see Fig. \ref{fig:framework}). Unlike previous approaches, that rely on training CNNs \cite{robinson2018real} or computationally expensive atlas-based registration methods as reverse classifiers \cite{rca,robinson2017automatic}, in-context learning models are computationally efficient and can adapt their predictions based on a small support set. 
We also introduce a dynamic selection of the reference dataset on a per-case basis based on retrieval similarity, enabling the use of smaller, more relevant reference datasets. This approach improves both computational efficiency and prediction quality. Implementation details, the specific in-context models explored (UniverSeg, SAM 2), and the retrieval-augmented experiments are provided in Section~\ref{sec:experimental-setup}. 
\\

\begin{figure*}[t]
    \centering
    \includegraphics[width=0.9\textwidth]{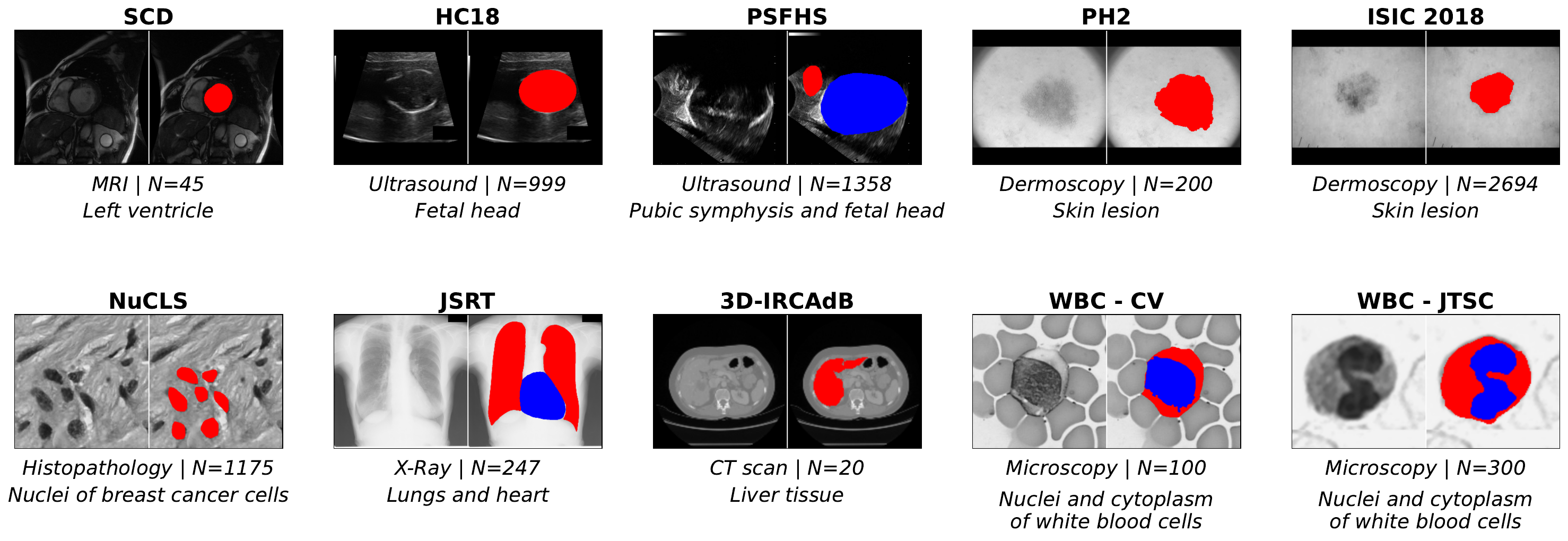} 
    \caption{Datasets overview including modality, number of images and structure.}
    \label{fig:datasets}
\end{figure*}

\section{Conformal RCA}
\subsection{Preliminaries: Split Conformal Prediction}
\label{sec:cp-preliminaries}
Conformal Prediction (CP) is a model-agnostic, distribution-free framework for uncertainty quantification that provides statistically valid prediction sets or intervals, for classification and regression tasks respectively, under minimal assumptions \cite{algorithmic-learning,conformalprediction}. Given a desired coverage level $1 - \alpha$ with $\alpha \in (0,1)$ and a ``nonconformity score'' function, CP constructs a prediction region $C_\alpha(X)$ for a given sample $X$ such that $Y \in C_\alpha(X)$ with probability $1-\alpha$, where $Y$ is the true label. Here, the desired coverage level $1-\alpha$ specifies the proportion of times the true outcome is expected to fall within the prediction set, while the nonconformity score quantifies how unusual or “nonconforming” a candidate prediction is with respect to the observed data. 

We focus on \emph{split conformal prediction}, a practical variant that conformalizes a fixed model using a held-out calibration set. This contrasts with \emph{full conformal prediction}, which typically requires retraining the model for each new test point. 
The only required assumption is that the calibration samples and the test samples are exchangeable (e.g., the samples are i.i.d.). Let us now state the general procedure for split conformal prediction, as presented in \cite{conformalprediction}. Given a calibration set $\{(x_i, y_i)\}_{i=1}^n$ and a new test input $x_{n+1}$:

\vspace{0.5em}
\begin{enumerate}[label=\arabic*.]
    \item Define a nonconformity score function $s(x, y) \in \mathbb{R}$ that quantifies the disagreement between the model's prediction for input $x$ and a ground-truth label $y$ (e.g. the absolute difference between the predicted and expected DSC score). Higher scores indicate poorer fit or higher error.
    \item Compute $\hat{q}$ as the $\lceil(1 - \alpha)(1+1/n)\rceil$-th empirical quantile of the calibration scores $\{s(x_1, y_1), \ldots, s(x_n, y_n)\}$. For example, for a desired coverage of 90\% ($\alpha = 10\%$), $\hat{q}$ will be the nonconformity score under which 90\% of the nonconformity scores computed in the calibration set fall.
    \item Construct the prediction set for $x_{n+1}$ as: 
    $$C_\alpha(x_{n+1}) = \{y : s(x_{n+1}, y) \leq \hat{q}\}.$$
    This set (or interval in case of regression problems) will contain all candidate outputs for $x_{n+1}$ that have a nonconformity score less than or equal to the threshold.
\end{enumerate}
This procedure gives the following coverage guarantee:

\begin{theorem}
\textit{Suppose that $\{(X_i, Y_i)\}_{i=1}^n$ and $(X_{n+1}, Y_{n+1})$ are i.i.d. Let $\hat{q}$ and $C_\alpha(X_{n+1})$ be as defined above. Then the following holds:}
\begin{equation}
    \mathbb{P}(Y_{n+1} \in C_\alpha(X_{n+1})) \geq 1 - \alpha.
    \label{thm:marginal-coverage}
\end{equation}
\end{theorem}

\vspace{0.5em}
The proof of this result traces back to \cite{algorithmic-learning}. Although the coverage guarantee holds regardless of the score function, the usefulness of the prediction sets is determined by how well it captures the model’s uncertainty. In practice, conformal prediction methods are primarily evaluated in terms of \textit{validity} (i.e., achieving the desired coverage level) and \textit{efficiency} (i.e., producing prediction sets or intervals that are as tight as possible). 

In regression problems, a widely adopted nonconformity score is the residual $s(x, y) = |y - \hat{f}(x)|$, where $\hat{f}(x)$ denotes the model’s prediction for input $x$ \cite{algorithmic-learning}. The resulting conformal prediction interval takes the form $C_\alpha(x_{n+1}) = \hat{f}(x_{n+1}) \pm \hat{q}$. Note that this leads to prediction intervals of equal width for all test inputs, regardless of their location in the input space. To better account for heteroscedasticity, i.e. when the noise variance varies across the input domain, a locally weighted nonconformity score has been proposed: $s(x, y) = \frac{|y - \hat{f}(x)|}{\hat{\sigma}(x)}$, where $\hat{\sigma}(x)$ is the estimated standard error of the residual \cite{lei2017distributionfreepredictiveinferenceregression}. 

An alternative and popular approach that naturally accounts for heteroscedasticity is Conformalized Quantile Regression (CQR) \cite{cqr}. CQR, builds on quantile regression models, which are trained to predict the conditional lower and upper quantiles $ \hat{t}_{\alpha/2}(x)$ and $\hat{t}_{1 - \alpha/2}(x)$ of the response $Y$ given an input \\ $X=x$. The nonconformity score is defined as the distance from $y$ to the nearest quantile:
\begin{equation*}
    s(x, y) = \max\left\{ \hat{t}_{\alpha/2}(x) - y,\; y - \hat{t}_{1 - \alpha/2}(x)\right\}
\end{equation*}
and the prediction interval for a new input $x_{n+1}$ is constructed as:
\begin{equation*}
    C_\alpha(x_{n+1}) = \left[ \hat{t}_{\alpha/2}(x_{n+1}) - \hat{q},\; \hat{t}_{1 - \alpha/2}(x_{n+1}) + \hat{q} \right].
\end{equation*}
We build on the CQR framework as the basis for our proposed method. This approach maintains the coverage guarantee of CP while often producing tighter, more locally adaptive intervals compared to residual-based methods, particularly in heteroscedastic settings \cite{cqr}.

\subsection{Proposed Method}
\label{sec:conformal-RCA-method}
Traditional RCA yields a point estimate of segmentation quality for an individual image, as defined in Eq.~\eqref{eq:RCA-estimate}. However, it does not quantify the uncertainty associated with this prediction. To address this limitation, we extend RCA following a split conformal prediction approach, producing prediction intervals with statistical guarantees under standard conformal assumptions.

Let $\mathcal{C} = \{(I_j^C, S_{I_j^C}, S_{I_j^C}^{GT})\}_{j=1}^n$ be a calibration set of images with both predicted ($S_{I_j^C}$) and ground-truth ($S_{I_j^C}^{GT}$) segmentations. For each calibration sample $(I_j^C, S_{I_j^C})$, we apply the RCA procedure: we train a reverse segmenter $f_{I_j^C, S_{I_j}^C}$ using $S_{I_j^C}$ as pseudo-ground truth, evaluate it on the reference database $D$, and obtain the corresponding RCA scores $\mathcal{Y}_j^C$ following Eq.~\eqref{eq:RCA-scores}. We also compute the true quality score $y(S_{I_j^C}) = \rho(S_{I_j^C}, S_{I_j^C}^{GT})$ of each sample in the calibration set.

To construct the prediction intervals, we follow an approach inspired by CQR but differing in a key aspect: instead of training a quantile regression model, we compute empirical quantiles directly from the RCA score distribution $\mathcal{Y}$. This allows us to naturally capture the uncertainty inherent in the RCA procedure while maintaining computational efficiency.

From each RCA score distribution $\mathcal{Y}_j^C$, we compute empirical quantiles at asymmetric levels that are chosen to reflect RCA's characteristic behavior. Specifically, the lower quantile $q_{\ell}(S_{I_j^C}) = \text{Quantile}(\mathcal{Y}_j^C, p_{\ell})$ is selected near the center of the distribution, while the upper quantile $q_h(S_{I_j^C}) = \text{Quantile}(\mathcal{Y}_j^C, p_h)$ is chosen closer to the maximum. This asymmetry captures the observation that RCA tends to underestimate segmentation quality, making higher scores more informative \cite{rca}.

We then define nonconformity scores that measure how far the true quality deviates from the RCA-based interval bounds:
\begin{equation}
    s_j = \max\{q_{\ell}(S_{I_j^C}) - y(S_{I_j^C}); \ y(S_{I_j^C}) - q_h(S_{I_j^C})\}.
    \label{eq:conformal-scores}
\end{equation}
Based on the calibration nonconformity scores, $\{s_j\}_{j=1}^n$, the ``CP threshold'', $\hat{q}$, is computed as the $\lceil(1 - \alpha)(1 + 1/n)\rceil$-th smallest value among $\{s_j\}_{j=1}^n$ (see Step 2 in Section \ref{sec:cp-preliminaries}). For a new test pair $(I, S_I)$, we compute its RCA scores $\mathcal{Y}$ and compute the corresponding empirical quantiles $q_\ell(S_I)$ and $q_h(S_I)$. The conformalized prediction interval is then constructed as: 
\begin{equation}
    C(S_I) = \left[q_{\ell}(S_I) - \hat{q}, q_h(S_I) + \hat{q}\right].
    \label{eq:interval}
\end{equation}

Assuming exchangeability between the test sample and calibration set, the constructed interval provides the marginal coverage guarantee stated in Theorem \ref{thm:marginal-coverage}, as shown in \cite{cqr}. In our experiments, $\rho$ corresponds to the Dice Similarity Coefficient (DSC), which takes values in the $[0, 1]$ range. Accordingly, we clip the predicted interval $C(S_I)$ to this range to ensure plausible quality estimates.
Note that the procedure can be applied with any reverse segmenter and can thus be readily combined with the proposed in-context approach.

\section{Experiments and results}
\label{sec:experiments}

\subsection{Datasets \& Computing Resources}
We evaluate the proposed Conformal and In-Context RCA variants across diverse datasets listed in Fig. \ref{fig:datasets} including SCD \cite{scd}, HC18 \cite{hc18}, PSFHS \cite{psfhs}, PH2 \cite{ph2}, ISIC 2018 \cite{isic2018}, NuCLS \cite{nucls}, JSRT \cite{jsrt}, 3D-IRCAdB \cite{3d-ircadb}, WBC-CV and WBC-JTSC \cite{wbc}. Unlike previous studies that explored RCA focusing on cardiac MRI \cite{robinson2017automatic,robinson2018real} or multi-organ segmentation on whole-body MRI \cite{rca} only, we broaden our analysis to a wider range of medical imaging challenges.
To maintain consistency, we convert all images to grayscale and resize them to 256x256 pixels. For 3D images, we extract slices, keeping those where the region of interest (ROI) is present.

All the experiments that will be described here were conducted on an AMD Ryzen 7 4800H CPU (8 cores) and an NVIDIA GTX 1660 Ti GPU (6GB of VRAM).

\begin{figure*}[t]
    \centering
    \includegraphics[width=\textwidth]{figures/cosar3.pdf} 
    \caption{Scatter plots comparing predicted vs real DSC on 3D-IRCAdB (\textbf{Left}) and JSRT (\textbf{Right}) for different reference datasets of size $k$ using UniverSeg as the reverse classifier. \textbf{Top}: results using randomly selected reference samples. \textbf{Bottom}: results following retrieval-augmentation, where the $k$ most similar images are selected based on cosine similarity in the DINOv2 embedding space. In both cases retrieval-augmentation achieves better performance with fewer samples, as shown by higher correlation and lower MAE (bold indicates the best value).}
    \label{fig:RA-comparison}
\end{figure*}

\begin{figure*}[t]
    \centering
    \includegraphics[width=\textwidth]{figures/cosar4.pdf} 
    \caption{Scatter plots comparing the predicted DSC across all datasets for In-Context RCA using UniverSeg and SAM 2 as reverse classifiers, followed by the retrieval-augmented and traditional Atlas RCA versions. All evaluations were conducted with a reference dataset of size 8. Bold values indicate the best performing methods exhibiting significant differences with respect to the non-bold values ($p < 0.01$, paired test), but no significant differences among themselves. In all cases 95\% confidence intervals obtained via bootstrapping.}
    \label{fig:DSC-all}
\end{figure*}

\subsection{Experimental Setup}
\label{sec:experimental-setup}
We evaluate In-Context RCA using two representative in-context segmentation models:

\begin{itemize}
    \item \textbf{UniverSeg}: a UNet-like architecture trained on a large collection of medical segmentation datasets (MegaMedical) and designed to generalize to unseen medical segmentation tasks without additional training.
    \item \textbf{SAM 2}: a foundational segmentation model originally developed for general-domain image and video segmentation, recently adapted to few-shot medical image segmentation \mbox{\cite{fsmedsam2}}.
\end{itemize}

For both models, we use the predicted segmentation under evaluation as the support mask and apply the model in a reverse-classification manner to segment reference images following the RCA protocol. \\

\noindent \textbf{UniverSeg-based reverse segmenter.} The UniverSeg architecture \cite{universeg} was designed to tackle unseen medical segmentation tasks without requiring additional training. Given a query image $Q$ and a small support set $\mathcal{S}$ of image-label pairs that define a new segmentation task, UniverSeg is able to segment the query image $Q$ following the provided examples, without updating the neural weights. It follows a UNet-like encoder-decoder architecture composed of CrossBlock modules. These modules, which are convolutional layers, enable the transfer of relevant information between the support set and the query image, allowing the model to perform accurate segmentation without task-specific retraining. It was trained on a collection of 53 open-access medical segmentation datasets called MegaMedical, covering a wide range of image modalities and structures. Here we adopt UniverSeg as the reverse classifier of the In-Context RCA framework. To this end, we use the target image $I$ and the predicted segmentation under assessment $S_I$ as the UniverSeg support set, i.e., we simply define $\mathcal{S}=\{(I, S_I)\}$. We then segment all the reference images $J_i$ in our RCA reference database by considering them as query images, one at a time, and finally compare each resulting segmentation $S_{J_k}$ with the corresponding reference segmentation $S_{J_k}^{GT}$ to estimate $\hat{\rho}(S_I)$ as defined in Eq.~\eqref{eq:RCA-estimate}. \\

\noindent \textbf{SAM 2-based reverse segmenter.} Segment Anything Model (SAM) \cite{sam} is a foundation model for image and video segmentation guided by prompts. Its architecture consists of image and prompt encoders that extract features, and a mask decoder that fuses this information to generate segmentations. SAM 2 \cite{sam2} extends its predecessor by incorporating memory components (an encoder, bank, and attention mechanism) enabling adaptation to new tasks and outperforming existing video object segmentation models, especially for temporal tracking. For medical few-shot segmentation (FSS) \cite{fsmedsam2}, SAM 2 treats the support set as initial frames of a sequence, with their segmentation masks serving as prompts. Despite having no temporal connection between images, this approach effectively transfers segmentation masks from annotated to unannotated examples (see \cite{fsmedsam2} for more details). In our framework, we use SAM 2 as a reverse classifier, following the few-shot setting with sequences that comprise only two frames: 1) the target image and its pseudo-ground truth, and 2) a query image drawn from the reference dataset. \\

\noindent \textbf{Retrieval-augmented  (RA) reference selection.} In order to further improve the efficiency of RCA, we propose to reduce the size of the RCA reference database by dynamically retrieving the most relevant reference images on a per-case basis. This allows the model to focus on smaller, more relevant subsets of data, optimizing computational efficiency while maintaining and even improving the quality of predictions. To implement this, we embed all the images from the reference database using DINOv2 embeddings \mbox{\cite{dinov2}} and then create a vector database using FAISS \mbox{\cite{faiss}}, which enables efficient similarity search. For each incoming target image whose segmentation quality we want to assess, we compute its embedding and retrieve the $k$ most similar images from the reference index based on cosine similarity. While DINOv2 was not explicitly trained for the medical domain, it has shown strong generalization, making it effective for general tasks, including those in medical imaging. In fact, a similar retrieval-augmented approach for few-shot medical image segmentation has been explored in \mbox{\cite{rag-sam2}}, demonstrating its effectiveness in similar scenarios. In the case of X-ray images, we leverage RAD-DINO \mbox{\cite{rad-dino}}, a domain-specific version of DINOv2 fine-tuned for radiology, as we expect it to further improve performance.

\subsection{Evaluation of In-Context RCA}
\noindent \textbf{Simulating masks of diverse quality.} To evaluate In-Context RCA, we generate segmentation masks of varying quality by training a small U-Net \cite{unet} for 10 epochs and saving the lower and higher-quality segmentations. Our goal is to achieve a uniform distribution of mask qualities with varying DSC for a comprehensive evaluation.\\


\noindent \textbf{Impact of dataset size and RA on performance.} We evaluate the impact of the reference dataset size and the use of RA (retrieval augmentation) by applying In-Context RCA with UniverSeg as our reverse classifier, considering different support set sizes. We aim to determine if RA maintains or even improves prediction quality while using a smaller reference dataset, thus reducing inference time in practice. We considered reference dataset sizes of 2, 5, 10 and 25 examples. 
Larger sets noticeably increase inference time per image without any meaningful improvement in prediction accuracy. 
Since UniverSeg supports only binary segmentation, for multi-class we process them separately (including background), and then select the most likely class per pixel using softmax, as described in \cite{universeg}.

Fig. \ref{fig:RA-comparison} shows the results on 3D-IRCAdB and JSRT datasets, comparing the performance of In-Context RCA for different reference datasets of size $k$. The figure contrasts two selection methods: random selection (first row) and RA (second row), where the $k$ most similar images are chosen based on cosine similarity in the DINOv2 embedding space. 
The horizontal axis represents the predicted DSC, while the vertical axis shows the real value. As the number of reference samples increases, both methods show improved performance, yielding higher correlation and lower mean absolute error (MAE). However, RA achieves better results with fewer reference samples, what is particularly beneficial during deployment, as it reduces inference time. Other similarity metrics, such as Euclidean distance and inner-product were also considered, but cosine similarity showed the best results. 

For a more complete evaluation, including MedImageInsight \cite{medimageinsight} embeddings pretrained on diverse medical imaging modalities, as well as statistical testing and confidence intervals estimated via bootstrapping, see Appendix.
\\

\begin{figure}[hbtp]
    \centering
    \includegraphics[width=\columnwidth]{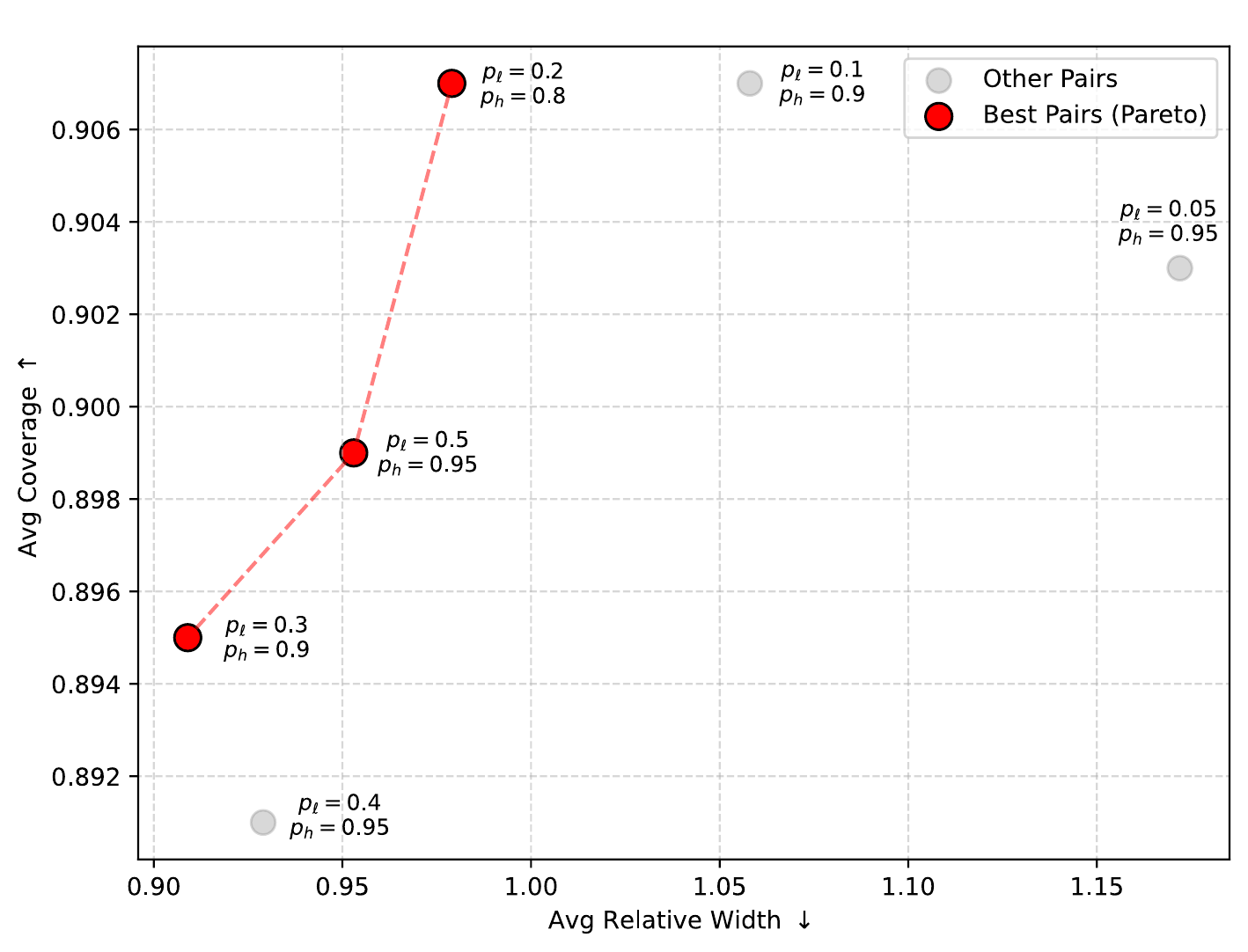} 
    \caption{Sensitivity analysis of conformal quantile selection for Conformal-RCA using SAM~2 as the reverse segmenter. Each point corresponds to a $(p_{\ell}, p_h)$ quantile pair, evaluated by its average empirical coverage and average relative prediction interval width across all datasets. Interval widths are normalized per dataset to ensure equal weighting across tasks. The Pareto frontier highlights quantile pairs offering optimal coverage–efficiency trade-offs.}
    \label{fig:quantile-pareto}
\end{figure}

\noindent \textbf{Comprehensive evaluation across datasets.} In order to assess the effectiveness and limitations of In-Context RCA, we conducted an extensive evaluation using UniverSeg and SAM 2 as reverse classifiers, considering DSC as evaluation metric.
We compared the performance of In-Context RCA with a single-atlas approach, known to offer the strongest results in the classical RCA framework \cite{rca,robinson2017automatic}. For the atlas, we included both traditional and RA versions, whereas for UniverSeg and SAM 2, we focused exclusively on the RA version as it showed better pefromance in our previous experiment (see Fig. \ref{fig:RA-comparison}). We used the segmentation masks generated by a UNet trained for different epochs as previously discussed, to ensure evaluation across varying segmentation qualities. We observe that following RA reference sample selection yields good results without requiring a large reference database. In all cases, we fixed the size of the reference dataset to 8, based on previous observations. 

Fig. \ref{fig:DSC-all} provides a comprehensive overview of the results when predicting the DSC across all datasets for all four variants: the traditional RCA framework using single-atlas, its RA version, and the In-Context RCA methods using UniverSeg and SAM 2 as reverse classifiers. We observe that In-Context RCA demonstrates similar and, in some cases, better results compared to the Atlas method, while being significantly more computationally efficient. Particularly in highly irregular segmentation problems like the NuCLS histopathology dataset, our SAM 2 variant achieves good correlation of about 0.76 while Atlas methods completely fail.

For additional results on representative datasets, including correlation to ASSD and HD95 estimations as well as comparisons to image-level quality estimation derived from pixel-level uncertainty methods such as MC-Dropout and TTA, see Appendix.
\\

\noindent \textbf{Performance gains for In-Context RCA.}
Importantly, while the classic RCA-atlas method takes on average one minute to process each image, SAM 2 and UniverSeg achieve the same task in just 0.70 seconds and 0.37 seconds respectively, approximately 85 and 162 times faster. This efficiency, combined with its strong performance, makes In-Context RCA an ideal choice for integration into clinical pipelines. 


\begin{table*}[htbp]
\centering
\caption{Dataset sizes and achieved coverage. For each dataset, we report the number of test and calibration samples and the average coverage (\%) achieved by each method.}
\label{tab:dataset-sizes}
\begin{tabular}{lrrrrrr}
\toprule
Dataset & SAM 2 & UniverSeg & Atlas-RA & PHISeg & PHISeg (CQR) & $N_\text{test}/N_\text{cal}$ \\
\midrule
3D-IRCAdB Liver      & 90.5 & 94.4 & 89.3 & 97.6 & 99.6 & 252 / 273 \\
HC18                 & 87.1 & 90.0 & 89.3 & 89.6 & 91.7 & 839 / 815 \\
ISIC 2018            & 93.5 & 93.5 & 91.0 & 85.4 & 92.9 & 479 / 482 \\
JSRT                 & 87.1 & 93.2 & 91.3 & 91.7 & 89.9 & 375 / 360 \\
NuCLS                & 91.0 & 91.8 & 90.8 & 92.8 & 93.6 & 391 / 385 \\
PH2                  & 82.1 & 95.5 & 96.1 & 92.7 & 94.4 & 179 / 177 \\
PSFHS                & 86.1 & 88.2 & 88.3 & 92.6 & 92.6 & 598 / 598 \\
SCD                  & 90.7 & 87.6 & 94.6 & 91.5 & 91.1 & 258 / 236 \\
WBC CV               & 94.8 & 90.5 & 93.8 & 80.7 & 85.9 & 153 / 151 \\
WBC JTSC             & 90.9 & 85.1 & 82.0 & 84.4 & 86.2 & 225 / 225 \\
\bottomrule
\end{tabular}
\end{table*}

\begin{figure*}[t]
    \centering
    \includegraphics[width=0.95\textwidth]{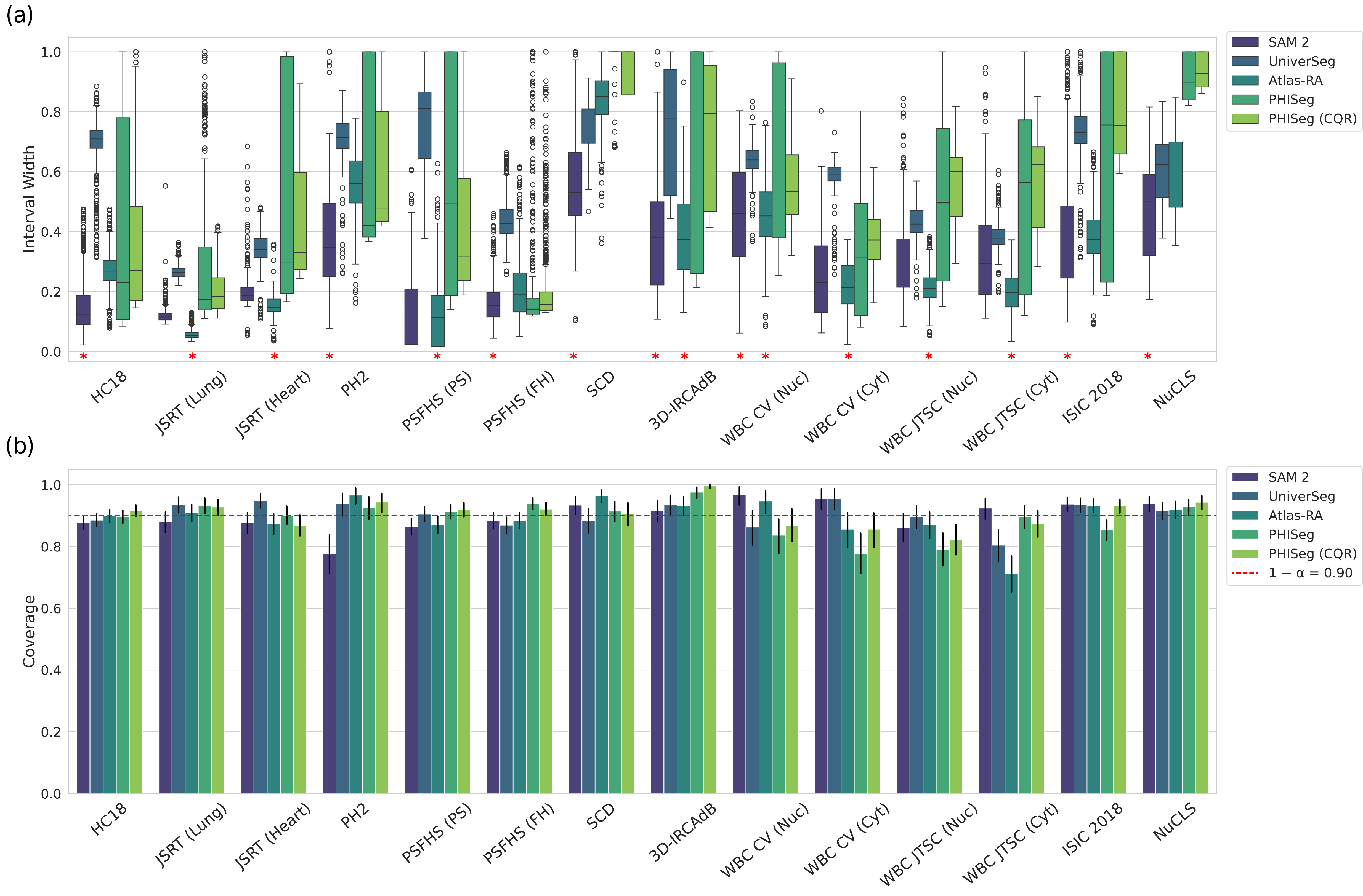} 
    \caption {(\textbf{a}) Distribution of prediction interval sizes (the lower the better) and (\textbf{b}) empirical coverage rates (the larger the better) for conformal prediction methods across all evaluated datasets. Results compare our Conformal-RCA variants using different reverse segmenters (Atlas-based, UniverSeg, SAM 2) against the baseline method PHiSeg from \cite{range-prediction}, as well as our proposed variant PHiSeg (CQR), which follows the empirical CQR approach. Target coverage level is set to 90\% ($\alpha = 0.1$). Asterisks in panel (\textbf{a}) indicate the best performing methods exhibiting significant differences with respect to the other methods for the same dataset ($p < 0.01$, paired Wilcoxon test), but no significant differences among themselves. Black vertical lines in panel (\textbf{b}) show 95\% confidence intervals obtained via bootstrapping.}
    \label{fig:conformal-results}
\end{figure*}

\subsection{Evaluation of Conformal RCA}
\label{sec:conformal-rca results}
To evaluate our conformal RCA approach, we split each dataset into training (80\%), calibration (10\%) and test (10\%) subsets (see Table~\ref{tab:dataset-sizes} for dataset sizes). The calibration set is used to compute the CP threshold $\hat{q}$ that adapts the prediction intervals, while performance is reported on the held-out test set.

In order to enable direct comparison with \cite{range-prediction}, we adopt PHiSeg \cite{phiseg}---the top performer in that work---as our backbone segmentation model. Importantly, Conformal-RCA remains model-agnostic, making it applicable to any underlying segmentation network regardless of output format. Segmentation masks of varying quality for both calibration and test sets were generated by training PHiSeg and saving intermediate checkpoints, following the same procedure as before. For PHiSeg, we follow the conformal calibration procedure described in \cite{range-prediction}: for each input image, $N = 50$ stochastic segmentations are generated, and the DSC is estimated from the softmax outputs \cite{li2022estimatingmodelperformancedomain}, yielding a distribution of predicted DSCs per sample. Nonconformity scores are defined as the absolute residuals between the mean predicted and ground truth DSC, scaled by the empirical standard deviation of the distribution. We additionally explore PHiSeg (CQR), a newly proposed variant where nonconformity scores are computed following the empirical CQR method. In this case, empirical quantiles are computed at levels 0.2 and 0.8.

Fig. \ref{fig:conformal-results} reports the distribution of prediction interval sizes and empirical coverage levels across all datasets for the evaluated methods. We include different versions of our approach, which differ only in the choice of reverse segmenter: the classical atlas-based method, and our two proposed in-context learning-based alternatives—UniverSeg and SAM 2. For all variants, we used a reference database of size 32, selected following the RA approach previously described. Empirical quantiles were computed at levels $p_{\ell} = 0.3$ and $p_h = 0.9$ and prediction intervals were constructed with a user-specified coverage of 90\% ($\alpha = 0.1$) as described in Section \ref{sec:conformal-RCA-method}. These values were selected based on a systematic sensitivity analysis over multiple symmetric and asymmetric quantile pairs, evaluating the trade-off between empirical coverage and interval width across datasets. The resulting Pareto frontier analysis, shown in Fig.~\ref{fig:quantile-pareto}, indicates that the chosen levels provide a stable balance between coverage and interval tightness.

Panel (a) of Fig. \ref{fig:conformal-results} shows the distribution of prediction interval widths across all datasets. Among our Conformal-RCA variants, SAM 2 and Atlas achieve consistently tight intervals, whereas UniverSeg yields somewhat wider intervals. Notably, SAM 2 offers a substantial computational speed‑up over the atlas‑based approach while maintaining similar interval quality.  Compared to the PHiSeg baseline, UniverSeg occasionally produces wider intervals; however, both SAM 2 and Atlas consistently attain lower median widths and reduced spread, indicating more robust uncertainty quantification. Importantly, compared to the conformal procedure used in prior work \cite{range-prediction}, the CQR variant of PHiSeg produces tighter intervals, further highlighting the benefits of following our proposed empirical CQR method within the CP framework.

As shown in panel (b) of Fig. \ref{fig:conformal-results}, neither our Conformal‑RCA nor the PHiSeg variants consistently reach the target marginal coverage of 90\%. This under-coverage phenomenon was also observed in \cite{range-prediction} and is likely due to violations of the exchangeability assumption between the test and calibration splits, caused by domain shifts and heterogeneous data sources. To further support this hypothesis, we ran controlled experiments in which both calibration and test sets are drawn from the same known distribution. In these controlled settings where the exchangeability assumption holds, the desired nominal coverage level of 90\% is achieved (see Appendix).

We further analyze the effect of calibration set size on conformal performance by varying the number of calibration samples for a fixed dataset and method. Results in Table~\ref{tab:calibration-size} show that coverage improves as calibration size increases. Additional results on representative datasets, showing intervals and coverage for ASSD and HD95 metrics as well as comparisons to MC-Dropout and TTA can be found in the Appendix.

\section{Conclusion}
In this work, we addressed the critical challenge of evaluating medical image segmentation quality in the absence of ground truth by presenting two complementary frameworks. First, In-Context RCA leverages the power of in-context segmentation models as reverse classifiers and integrates retrieval augmentation to select relevant reference samples. This combination enables efficient quality estimation using smaller reference databases while either matching or improving prediction accuracy of traditional RCA.

Second, Conformal RCA extends the traditional RCA procedure using split conformal prediction to produce prediction intervals with theoretical statistical guarantees under standard assumptions, providing a more comprehensive assessment of segmentation quality that moves beyond point estimates. Unlike previous interval-estimation approaches that rely on specialized probabilistic architectures and require calibrated softmax outputs, Conformal RCA is fully model‑agnostic and easily integrates with any segmentation network.

Experimental results conducted across diverse medical imaging modalities, show that In-Context RCA achieves comparable or superior performance to traditional atlas-based approaches  while being up to two orders of magnitude faster. It also overcomes the limitations imposed by atlas-based methods allowing to tackle scenarios beyond anatomical segmentation, such as cell images. Conformal RCA demonstrates the ability to produce meaningful prediction intervals that outperform existing conformal prediction approaches for segmentation quality control.

A remaining limitation is the reliance of exchangeability between calibration and test sets. When this assumption is violated (e.g., due to domain shifts in scanner settings or patient populations), coverage can degrade which may affect reliability in real-world clinical deployments. Developing methods that are robust to distribution shifts is an important direction for future work, and remains an open problem even beyond medical imaging, as current approaches suffer from inexact adaptations and increased prediction sets \cite{clusteredcp,barber2023conformal}. In addition, validation on real deployment data and human-in-the-loop studies would further help establish clinical reliability and practical utility.

Overall, our results position both In-Context RCA and Conformal RCA as promising promising steps toward practical automated quality control in clinical workflows, where rapid and reliable segmentation assessment is essential.

\bibliography{tmi.bib}
\bibliographystyle{IEEEtran}

\newpage
\appendices
\section*{Appendix A: Additional Results}
\label{sec:additional-results}

\textcolor{black}{To further analyze the behavior of different quality-control approaches, we conduct additional experiments on 3D-IRCAdB and JSRT using newly trained segmentation models. Specifically, we train a UNet with dropout on each dataset and save intermediate checkpoints during training. These checkpoints are used to generate segmentation masks of varying quality, allowing us to evaluate how well different quality-estimation methods correlate with true segmentation performance. In this setting, we additionally compare Conformal RCA against two widely used uncertainty-based baselines: test-time augmentation (TTA) and MC Dropout. These methods estimate epistemic uncertainty from multiple stochastic segmentation outputs and are commonly used as proxies for segmentation quality.}

\subsection*{A.1 Correlation Analysis on New Metrics and Methods}
\textcolor{black}{We evaluate how well different quality-estimation approaches correlate with true segmentation performance measured by Dice similarity coefficient (DSC) as well as two distance-based metrics: average symmetric surface distance (ASSD) and 95th percentile Hausdorff distance (HD95). }

\textcolor{black}{
In-context RCA directly outputs a scalar quality estimate that is comparable to the target metric (e.g., predicted DSC or ASSD) on a per-image basis. In contrast, approaches like MC Dropout and TTA estimate epistemic uncertainty rather than segmentation quality itself. As a result, these methods require additional post-processing to derive an image-level quality proxy. Following \cite{uncertainty-maps}, we first compute the mean predictive probability map across $N=50$ output samples, then evaluate pixel-wise entropy for each class, and finally average entropy values to obtain a scalar uncertainty score. This score is subsequently correlated with the true segmentation metrics. Absolute correlation values are reported in Table~\ref{tab:metrics}.}

\subsection*{A.2 Conformal Results on New Metrics and Methods}
\textcolor{black}{For the same two datasets, we evaluate the performance of Conformal RCA in terms of prediction intervals and empirical coverage for DSC, ASSD and HD95. We compare our method against MC Dropout and TTA following \cite{range-prediction}. Results are shown in Fig. \ref{fig:conformal-additional}.}

\subsection*{A.3 Additional Retrieval-Augmentation Experiments}
\textcolor{black}{To broaden the analysis of retrieval-augmentation (RA) for Conformal RCA, we evaluate reference selection strategies using embeddings pretrained on different domains. For the previously mentioned datasets and with UniverSeg as the reverse classifier, we compare randomly selected reference sets to RA-based selection, where the $k$ most similar images are chosen based on embeddings from different backbone models: DINOv2 \cite{dinov2} (general domain), MedImageInsight \cite{medimageinsight} (pretrained on a large and diverse set of medical imaging modalities including X-ray, CT, MRI, dermoscopy, OCT, fundus photography, ultrasound, histopathology, and mammography), and, for JSRT, additionally RAD-DINO \cite{rad-dino} (X-ray specific). Results are shown in Fig. \ref{fig:ra-additional}.}

\textcolor{black}{Overall, RA consistently improves reference selection compared to random sampling. For 3D-IRCAdB, DINOv2 sometimes outperforms the domain-specific MedImageInsight embeddings, whereas for JSRT, RAD-DINO and MedImageInsight embeddings always outperform DINOv2, demonstrating the benefit of domain-adapted embeddings in radiological tasks.}

\begin{figure}[!t]
    \centering
    \includegraphics[width=\columnwidth]{figures/cosar7.pdf} 
    \caption{Predicted vs. true DSC for 3D-IRCAdB and JSRT using different reference dataset sizes and selection strategies. Retrieval-augmented selection is based on embeddings from DINOv2, MedImageInsight, or RAD-DINO (for JSRT). Bold indicates methods exhibiting significant differences with respect to others ($p < 0.01$, paired tests). 95\% confidence intervals were obtained via bootstrapping.}
    \label{fig:ra-additional}
\end{figure}

\begin{table}[t]
\centering
\caption{Absolute correlation to ASSD, Dice and HD95 metrics for different methods.
95\% confidence interval (CI) estimated via bootstrapping is reported in each case.}
\label{tab:metrics}
\resizebox{\columnwidth}{!}{%
\begin{tabular}{llcccc}
\toprule
Dataset & Metric & UniverSeg & SAM 2 & TTA & MC \\
\midrule
\multirow{3}{*}{3D-IRCADb}
& ASSD & {\boldmath$0.97 \pm 0.00$} & $0.96 \pm 0.01$ & $0.79 \pm 0.02$ & $0.68 \pm 0.02$ \\
& DSC  & $0.89 \pm 0.01$ & {\boldmath$0.97 \pm 0.01$} & $0.82 \pm 0.02$ & $0.77 \pm 0.01$ \\
& HD95 & {\boldmath$0.96 \pm 0.01$} & $0.95 \pm 0.01$ & $0.77 \pm 0.02$ & $0.73 \pm 0.02$ \\
\midrule
\multirow{3}{*}{JSRT (Lung)}
& ASSD & $0.91 \pm 0.02$ & {\boldmath$0.96 \pm 0.01$} & $0.83 \pm 0.04$ & $0.78 \pm 0.04$ \\
& DSC  & $0.85 \pm 0.05$ & {\boldmath$0.92 \pm 0.02$} & $0.79 \pm 0.05$ & $0.82 \pm 0.06$ \\
& HD95 & $0.79 \pm 0.04$ & {\boldmath$0.86 \pm 0.04$} & $0.66 \pm 0.07$ & $0.82 \pm 0.04$ \\
\midrule
\multirow{3}{*}{JSRT (Heart)}
& ASSD & {\boldmath$0.87 \pm 0.03$} & $0.83 \pm 0.03$ & $0.64 \pm 0.08$ & $0.13 \pm 0.05$ \\
& DSC  & {\boldmath$0.95 \pm 0.01$} & {\boldmath$0.95 \pm 0.01$} & $0.91 \pm 0.02$ & $0.54 \pm 0.09$ \\
& HD95 & {\boldmath$0.68 \pm 0.06$} & $0.48 \pm 0.10$ & $0.24 \pm 0.12$ & $0.11 \pm 0.06$ \\
\bottomrule
\end{tabular}
}
\end{table}

\begin{figure*}[!t]
    \centering
    \includegraphics[width=\textwidth]{figures/cosar8.pdf} 
    \caption{Distribution of prediction interval widths (\textbf{left}, lower is better) and marginal coverage (\textbf{right}, higher is better) for Conformal RCA across three metrics (DSC, ASSD, HD95) on the 3D-IRCAdB and JSRT datasets. Results compare different reverse segmenters (Atlas, UniverSeg, SAM 2) against uncertainty-based baselines (MC Dropout, TTA). Asterisks indicate a method exhibiting significant differences with respect to the others ($p < 0.01$, paired Wilcoxon test), while black vertical lines denote 95\% confidence intervals obtained via bootstrapping.}
    \label{fig:conformal-additional}
\end{figure*}

\subsection*{A.4 Effect of Calibration Set Size on Conformal Performance}
\textcolor{black}{We further analyze the effect of calibration set size on Conformal RCA performance. Using the 3D-IRCAdB dataset with SAM 2 as the reverse classifier, we vary the number of calibration samples $N_\text{cal}$ and report the resulting empirical coverage and average prediction interval width. Results are summarized in Table~\ref{tab:calibration-size}. Generally, coverage stabilizes as the calibration set size increases, highlighting the practical trade-offs involved when using smaller calibration sets.}

\begin{table}[htbp]
\centering
\caption{Effect of calibration set size on coverage and average prediction width for the 3D-IRCAdB dataset using SAM 2.}
\label{tab:calibration-size}
\begin{tabular}{ccc}
\toprule
\boldmath{$N_\text{cal}$} & \textbf{Coverage} & \textbf{Avg. Width} \\
\midrule
200  & 0.54 & 0.12 \\
400  & 0.55 & 0.13 \\
600  & 0.67 & 0.15 \\
800  & 0.81 & 0.19 \\
1000 & 0.78 & 0.18 \\
\bottomrule
\end{tabular}
\end{table}

\newpage
\section*{Appendix B: Controlled Experiments on Exchangeability Assumption}
\label{sec:controlled-experiments}

To illustrate how exchangeability underpins Conformal RCA’s coverage guarantees, we run a controlled experiment in which both calibration and test Dice scores are drawn from the same underlying distribution. RCA scores are then obtained by adding Gaussian noise to these true scores (clipped to [0,1]). We then apply Conformal RCA with the same settings used in Section \ref{sec:conformal-rca results}: a reference set of size 32 and empirical quantiles at levels $p_{\ell} = 0.3$ and $p_h = 0.9$.

\begin{figure}[!t]
    \centering
    \includegraphics[width=\columnwidth]{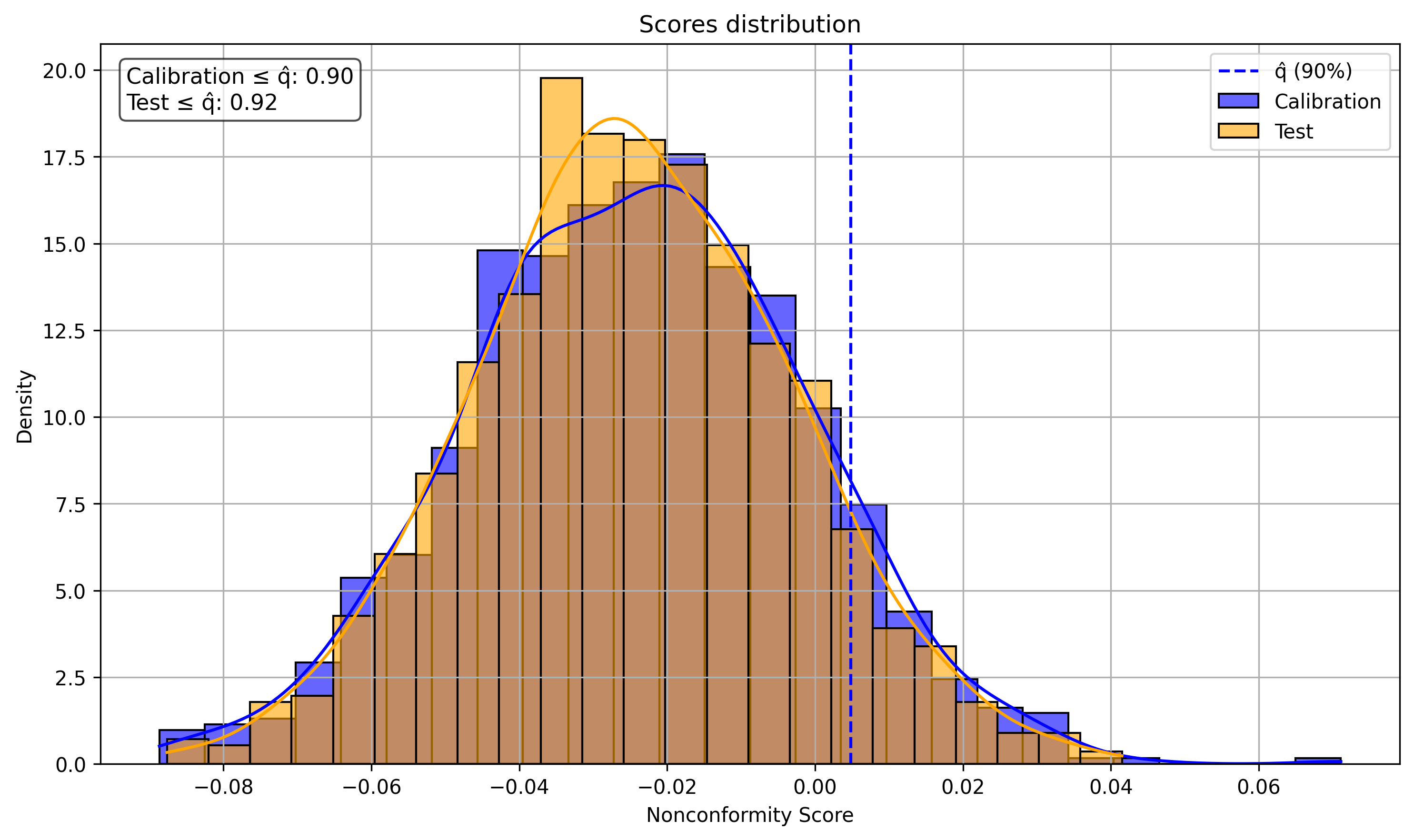} 
    \caption{Distribution of nonconformity scores for synthetic calibration and test sets, generated by sampling true Dice scores from a $\mathrm{Beta}(2, 2)$ distribution and adding noise drawn from \(\mathcal{N}(0, 0.1)\) to simulate variability in RCA scores. The vertical dashed line indicates the conformal quantile threshold \(\hat{q}\).}
    \label{fig:synthetic-scores}
\end{figure}

Fig. \ref{fig:synthetic-scores} shows the resulting nonconformity‑score histograms for calibration and test splits, along with the computed threshold $\hat{q}$. Empirical marginal coverage surpasses the desired 90 \% target level, confirming that Conformal RCA achieves valid coverage under ideal conditions.

\end{document}